\documentclass[twocolumn]{el-author}

\newcommand{\ua}{\uparrow}
\newcommand{\nc}{\newcommand}
\nc{\da}{\downarrow} \nc{\hc}{\hat{c}} \nc{\hS}{\hat{S}}
\nc{\bra}{\langle} \nc{\ket}{\rangle} \nc{\eq}{equation (\ref}
\nc{\h}{\hat} \nc{\hT}{\h{T}}\nc{\be}{\begin{eqnarray}}
\nc{\ee}{\end{eqnarray}}\nc{\rd}{\textrm{d}}\nc{\e}{eqnarray}\nc{\hR}{\hat{R}}\nc{\Tr}{\mathrm{Tr}}
\nc{\tS}{\tilde{S}}\nc{\tr}{\mathrm{tr}}\nc{\8}{\infty}\nc{\lgs}{\bra\ua,\phi|}\nc{\rgs}{|\ua,\phi\ket}
\nc{\hU}{\hat{U}}\nc{\lfs}{\bra\phi|}\nc{\rfs}{|\phi\ket}\nc{\hZ}{\hat{Z}}\nc{\hd}{\hat{d}}\nc{\mD}{\mathcal{D}}
\nc{\bd}{\bar{d}}\nc{\bc}{\bar{c}}\nc{\mc}{\mathcal}\nc{\ea}{eqnarray}\nc{\mG}{\mathcal{G}}\nc{\bce}{\begin{center}}
\nc{\ece}{\end{center}}
\date{12th December 2011}

\usepackage{amsmath}
\usepackage{amssymb,xspace,intmacros,subfigure,amsfonts,color}
\usepackage{epsfig,epsf,epstopdf,graphicx}

\newcommand{\Ib}{{\textbf{I}}}
\newcommand{\ub}{{\textbf{u}}}

\newcommand\ie{i.e.\xspace}

\begin{document}

\title{Weighted diffusion LMP algorithm for distributed estimation in non-uniform noise conditions}

\author{H. Zayyani, M. Korki}

\abstract{This letter presents an improved version of diffusion least mean p-power (LMP) algorithm for distributed estimation. Instead of sum of mean square errors, a weighted sum of mean square error is defined as the cost function for global and local cost functions of a network of sensors. The weight coefficients are updated by a simple steepest-descent recursion to minimize the error signal of the global and local adaptive algorithm. Simulation results show the advantages of the proposed weighted diffusion LMP over the diffusion LMP algorithm specially in the non-uniform noise conditions in a sensor network.}

\maketitle

\section{Introduction}

Distributed estimation is widely used in wireless sensor networks to estimate a parameter vector distributively and cooperatively \cite{Sayed14}. Among incremental \cite{Sayed14}, consensus \cite{Sayed14} and diffusion \cite{Sayed14}, \cite{LopS08}\nocite{CattS10}--\cite{Huan15} strategies for distributed estimation, in this letter, we focus on diffusion-based algorithms. A diffusion least mean square (LMS) algorithm has been proposed in \cite{LopS08} and \cite{CattS10}. Moreover, a diffusion least mean p-power (LMP) has been suggested in \cite{Wen13} for distributed estimation in alpha-stable noise environments. Also, a diffusion LMP algorithm with adaptive variable power has been proposed in \cite{Wen14}.

In this letter, the global and local cost functions of diffusion LMP algorithm are modified. The global cost function is defined as the weighted mean square error of all the sensor nodes. This is inspired by the non-uniform noise cases where some nodes in the sensor network operate under better noise condition. Hence, it is better to assign more weights to these nodes instead of uniform distribution of weightings among all nodes. For the local cost function, we consider a time varying combination coefficients or a time-varying weight instead of a constant combination coefficient or constant weight. The weights in the global and local cost functions are updated based on a steepest-descent recursion to minimize the mean square error of the adaptive algorithm.





\section{Problem formulation}
Consider a sensor network of $N$ nodes distributed over a region. Each sensor at time instant $n$ takes a scalar measurement $d_{k,n}$, which is a linear measurement of a common parameter vector $\boldsymbol{\omega}_o$. The model is
\begin{align}
  d_{k,n}=\boldsymbol{\omega}_o^{T}\ub_{k,n}+v_{k,n},
\end{align}
where $k$ is the sensor number, $\ub_{k,n}$ is the regression column vector, $v_{k,n}$ denotes the measurement noise and $T$ denotes the transposition. We aim to estimate the common parameter vector $\boldsymbol{\omega}_o$ based on linear measurements $d_{k,n}$ and knowing the regression vectors $\ub_{k,n}$. Similar to \cite{Wen13}, we assume that all the signals are real, and extension to complex case is straightforward. Each node can estimate the parameter vector $\boldsymbol{\omega}_o$ separately based on its own adaptive algorithm. However, in distributed estimation, we aim to cooperatively estimate the parameter vector $\boldsymbol{\omega}_o$ via in-network processing.

\section{The proposed weighted diffusion LMP algorithm}
For centralized global estimation of the diffusion LMP algorithm, the parameter vector $\boldsymbol{\omega}_o$ is estimated by minimizing the following global cost function \cite{Wen13}:
\begin{equation}
J_{\mathrm{LMP}}^{\mathrm{glob}}(\boldsymbol{\omega})=\sum_{k=1}^N\mathrm{E}\{|d_{k,n}-\boldsymbol{\omega}^T\ub_{k,n}|^p\},
\end{equation}
where $\mathrm{E}\{.\}$ is the expectation operator. Inspired by non-uniform noise conditions and the idea of combination of adaptive filters \cite{Aren16}, we propose to use the following global cost function for weighted diffusion LMP:
\begin{equation}
J_{\mathrm{WLMP}}^{\mathrm{glob}}(\boldsymbol{\omega})=\sum_{k=1}^N\alpha_k(n)\mathrm{E}\{|d_{k,n}-\boldsymbol{\omega}^T\ub_{k,n}|^p\},
\end{equation}
where $\alpha_k(n)$ is the adaptive weights for $k$'th sensor at time instant $n$ with the constraint $\sum_{k=1}^N\alpha_k(n)=1$. For the centralized estimation of the unknown parameter vector $\boldsymbol{\omega}$, a steepest-descent recursion is used, which is given by
\begin{equation}
\boldsymbol{\omega}_n=\boldsymbol{\omega}_{n-1}+\sum_{k=1}^N\mu_k\alpha_k(n)|e_{k,n}|^{p-2}e_{k,n}\ub_{k,n},
\end{equation}
where $e_{k,n}=d_{k,n}-\boldsymbol{\omega}^T_{k,n}\ub_{k,n}$ is the error signal. To update the weight coefficients $\alpha_k(n)$, similar to \cite{Aren16}, we assume that $\alpha_k(n)=\frac{\mathrm{e}^{a_k(n)}}{\sum_{k=1}^N\mathrm{e}^{a_k(n)}}$. We can update the coefficients $a_k(n)$ by a steepest-descent recursion to minimize the instantaneous error $e^2_{k,n}$. Therefore, we have:
\begin{equation}
a_k(n+1)=a_k(n)-\mu^{'}_a\frac{\partial e^2_{k,n}}{\partial a_k(n)},
\end{equation}
where final recursion of $\a_k(n+1)$ after some calculations is:
\begin{equation}
a_k(n+1)=a_k(n)-\mu_a\mu|e_{k,n}|^p\ub^T_{k,n}\ub_{k,n}b_{k,n},
\end{equation}
where $b_k(n)=\frac{\mathrm{e}^{a_k(n)}\sum_{k=1}^N\mathrm{e}^{a_k(n)}-(\mathrm{e}^{a_k(n)})^2 }{(\sum_{k=1}^N\mathrm{e}^{a_k(n)})^2}$.

For local cost function, we suggest to use a time-varying combination weight instead of the fixed combination weight. Therefore, the local cost function at $k$'th sensor is defined as:
\begin{equation}
J_{k}^{\mathrm{loc}}(\boldsymbol{\omega})=\sum_{l\in\mathbb{N}_k}c_{kl}(n)\mathrm{E}\{|d_{l,n}-\boldsymbol{\omega}^T\ub_{k,n}|^p\},
\end{equation}
where $c_{kl}$ is the combination weight from sensor $l$ to sensor $k$ with the constraint $\sum_{l=1}^N c_{kl}(n)=1$. Hence, similarly we can assume $c_{kl}(n)=\frac{\mathrm{e}^{a_{kl}(n)}}{\sum_{k=1}^N\mathrm{e}^{a_{kl}(n)}}$. Since the proposed weighted diffusion LMP has the same local cost function as diffusion LMP, the overall algorithm is the same except for updating the weight coefficient properly to reduce the estimation error. Hence, the overall algorithm is a three step algorithm. At the first step, intermediate estimates at each node is calculated by the following formula \cite{Wen13},
\begin{equation}
\boldsymbol{\phi}_{k,n-1}=\sum_{l\in\mathbb{N}_k}a_{1,kl}(n)\boldsymbol{\omega}_{l,n-1},
\end{equation}
where the coefficients $\{a_{1,lk}\}$ determine which nodes should share their intermediate estimates $\{\boldsymbol{\omega}_{l,n-1}\}$ with node $k$ \cite{Wen13}. At the second step, the nodes update their estimates by \cite{Wen13}
\begin{equation}
\boldsymbol{\psi}_{k,n}=\boldsymbol{\phi}_{k,n-1}+\mu_k\sum_{l\in\mathbb{N}_k}c_{kl}(n)|e_{l,n}|^{p-2}e_{l,n}\ub_{l,n}.
\end{equation}

Finally, at the third step, the second combination is performed as \cite{Wen13}
\begin{equation}
\label{eq: omega}
\boldsymbol{\omega}_{k,n}=\sum_{l\in\mathbb{N}_k}a_{2,kl}(n)\boldsymbol{\psi}_{l,n},
\end{equation}
where the coefficients $\{a_{2,lk}\}$ determine which nodes should share their intermediate estimates $\{\boldsymbol{\psi}_{l,n}\}$ with node $k$ \cite{Wen13}. For the simplicity, in the proposed weighted diffusion LMP, we assume that all the combination coefficients are equal, \ie $c_{kl}(n)=a_{1,kl}(n)=a_{2,kl}(n)=\frac{\mathrm{e}^{a_{kl}(n)}}{\sum_{k=1}^N\mathrm{e}^{a_{kl}(n)}}$. To update $c_{kl}(n)$, we update $a_{kl}(n)$ based on reducing the squared error $e^2_{k,n}$. To reduce this error, we use a steepest descent recursion as
\begin{equation}
a_{kl}(n+1)=a_{kl}(n)-\mu_a^{'}\frac{\partial e^2_{k,n}}{\partial a_{kl}(n)}\\
=a_{kl}(n)-\mu_a e_{k,n}\frac{\partial e_{k,n}}{\partial a_{kl}(n)}
\end{equation}
where $\mu_a=2\mu_a^{'}$. We also have $\frac{\partial e_{k,n}}{\partial a_{kl}(n)}=\frac{\partial e_{k,n}}{\partial c_{kl}(n)}\frac{\partial c_{kl}(n)}{\partial a_{kl}(n)}$. From $e_{k,n}=d_{k,n}-\boldsymbol{\omega}^T_{k,n}\ub_{k,n}$ and from (\ref{eq: omega}), noting that $a_{2,kl}(n)=c_{kl}(n)$ ,we have $\frac{\partial e_{k,n}}{\partial c_{kl}(n)}=-\boldsymbol{\psi}^T_{l,n}\ub_{k,n}$. We also have $\frac{\partial c_{kl}(n)}{\partial a_{kl}(n)}=d_{kl}(n)=\frac{\mathrm{e}^{a_{kl}(n)}\sum_{k=1}^N\mathrm{e}^{a_{kl}(n)}-(\mathrm{e}^{a_{kl}(n)})^2 }{(\sum_{k=1}^N\mathrm{e}^{a_{kl}(n)})^2}$. Hence, the overall recursion for updating $a_{kl}(n)$ is
\begin{equation}
a_{kl}(n+1)=a_{kl}(n)+\mu_a e_{k,n}\boldsymbol{\psi}^T_{l,n}\ub_{k,n}d_{kl}(n).
\end{equation}

\begin{figure}[h]
\hspace*{-1in}
\centering{\includegraphics[width=40mm, angle=-90]{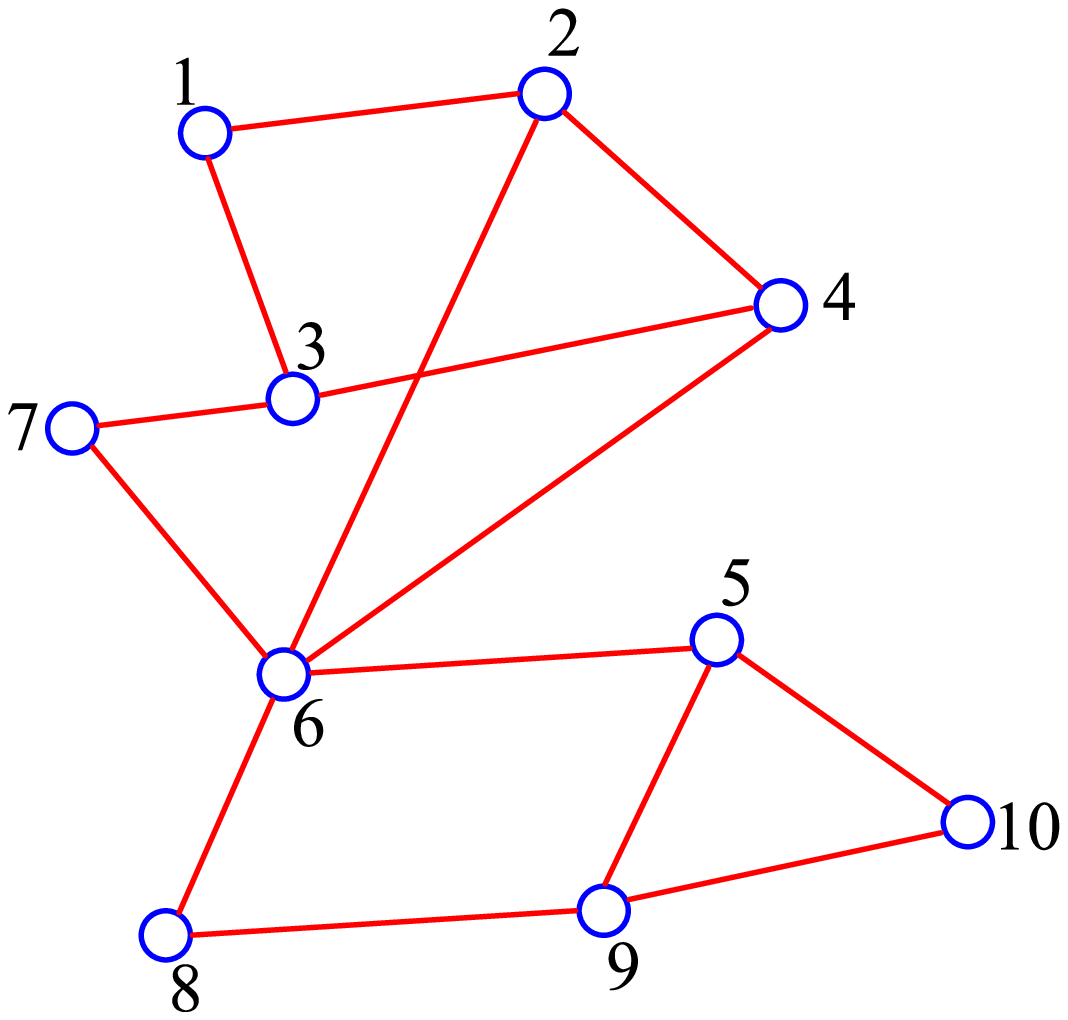}}
\vspace{5em}
\caption{Topology of the wireless sensor network with N=10 nodes.
\source{}}
\end{figure}

\begin{figure}[h]
\centering{\includegraphics[width=60mm]{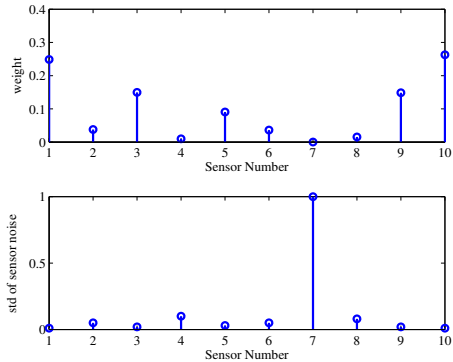}}
\caption{Non-uniform standard deviation (std) of Gaussian noise in sensors (bottom) and corresponding weights for the
proposed weighted diffusion LMP algorithm (top). Note that sensors with higher variance of noise have lower weights and vice versa.
\source{}}
\end{figure}

\begin{figure}[h]
\centering{\includegraphics[width=60mm]{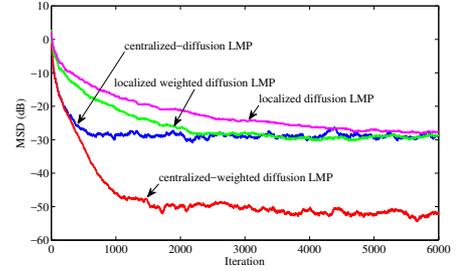}}
\caption{MSD versus iteration for different versions of diffusion LMP algorithm in Gaussian noise environments.
\source{}}
\end{figure}

\begin{figure}[h]
\centering{\includegraphics[width=60mm]{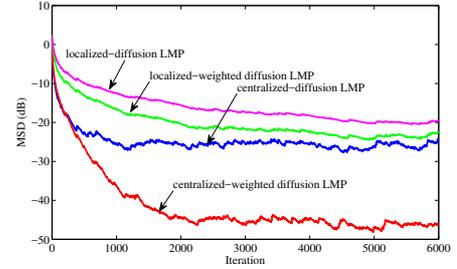}}
\caption{MSD versus iteration for different versions of diffusion LMP algorithm in alpha-stable noise environments.
\source{}}
\end{figure}

\section{Simulation results}

In our experiment, we consider a distributed network composed of 10 nodes (see Fig.~1). The size of the unknown vector parameter $\boldsymbol{\omega}_o$ is $M=50$. The vector elements are selected as a unit variance Gaussian distribution random variable. The measurement signal $\ub_{k,i}$ is a $1\times 50$ vector satisfying $\ub_{k,i}\sim \mathcal{N}\left ( 0,\sigma^2_{u,k}\Ib \right )$ with $\sigma_{u,k}=1$. We consider two cases for the measurement noise. At first case, the measurement noise $v_{k,i}$ is assumed to be Gaussian with zero mean and variance $\sigma^2_{n,i}$. The standard deviation (std) of noise in sensors is assumed to be non-uniform as depicted in Fig.~2. At the second case, the measurement noise $v_{k,i}$ is assumed to be impulsive. In wireless sensor networks (WSNs), the impulsive noise follows a symmetric alpha-stable distribution with the characteristic function $\phi(v_{k,i})=\exp(-\gamma|v_{k,i}|^\alpha)$ \cite{ZayyKM16}. The characteristic exponent $\alpha\in(0,2]$ controls the impulsiveness of the noise (smaller $\alpha$ leads to more frequent occurrence of impulses) and dispersion $\gamma>0$ describes the spread of the distribution around its location parameter which is zero for our purposes \cite{ZayyKM16}. The dispersion parameter $\gamma$ plays a similar role as the variance of Gaussian distribution \cite{Wen13}. We assume non-uniform dispersions for various sensors which are 0.01, 0.001, 0.02, 0.03, 0.002, 0.003, 0.02, 0.05, 0.005, and 0.1 for nodes 1 to 10, respectively. The exponent $\alpha$ is selected as 1.25. For performance metric, similar to \cite{LorS13}, we use mean square deviation (MSD) defined as $\mathrm{MSD}(dB)=20\mathrm{log}(||\boldsymbol{\omega}-\boldsymbol{\omega}_o||_2)$. The results are averaged over 50 independent trials. Figure~2 shows the standard deviation of noise in various sensors and the final learned weights for the proposed weighted diffusion LMP in Gaussian noise environments. It is seen that the weights for the sensors with higher variance of noise are lower than those for the sensors with lower variance of noise. Figure~3 shows MSD curves versus iteration index for 4 different algorithms in Gaussian noise environments. The algorithms are the centralized estimation of diffusion LMP, the centralized estimation of weighted diffusion LMP, the localized estimation of diffusion LMP and the localized estimation of the weighted diffusion LMP. For global centralized estimation, the global step size are selected as $\mu^{glob}=0.005$ and the other step size is selected as $\mu_a=10$. For the localized estimation, the local step size is selected equal to the global case \ie $\mu^{loc}=0.005$ and the other step size is selected as $\mu_a=0.01$. As it is shown in \cite{Wen13}, the diffusion LMP algorithm converges for the values of order $p$ close to 1, thus we set $p=1.2$ in all the simulations. It is seen that the proposed localized weighted diffusion LMP algorithm outperforms the localized diffusion LMP algorithm. Also, the proposed centralized-weighted diffusion LMP algorithm significantly outperforms the centralized-diffusion LMP algorithm. Figure 3 also shows that the best algorithm, compared to the others, is the centralized-weighted diffusion LMP algorithm. Figure~4 shows MSD curves versus iteration index for 4 different algorithms in the case of alpha-stable noise when all other parameters in the simulations remain unchanged. As it can be seen, the proposed centralized-weighted diffusion LMP algorithm still performs the best among all the other algorithms.

%

\section{Conclusion}
A weighted diffusion LMP algorithm has been proposed for distributed estimation in non-uniform noise environments. Unlike the diffusion LMP algorithm, which utilizes the uniform distribution of weights among sensors, the proposed weighted diffusion LMP algorithm assigns different weights to the sensors with different variance of noise to improve the performance. Compared with the diffusion LMP algorithm, better performance has been achieved for the proposed weighted diffusion LMP algorithm.


\vskip5pt

\noindent H. Zayyani (\textit{Department of Electrical and Computer Engineering, Qom University of Technology, Qom, Iran})
\vskip3pt

\noindent E-mail: zayyani@qut.ac.ir

\noindent M. Korki (\textit{School of Software and Electrical Engineering, Swinburne University of Technology, Hawthorn, Australia})
\vskip3pt

\noindent E-mail: mkorki@swin.edu.au

\end{document}